\documentclass{article}

\usepackage{times}
\usepackage{graphicx} 
\usepackage{subfigure} 

\usepackage{natbib}

\usepackage{algorithm}
\usepackage{algorithmic}
\usepackage{booktabs}
\usepackage{multirow}
\usepackage{paralist}
\usepackage{amsmath,bm}
\usepackage{amsfonts}
\usepackage{mathrsfs}
\usepackage{amsbsy,amsthm,amssymb}

\usepackage{hyperref}

\usepackage[accepted]{icml2017}

\icmltitlerunning{Coupling Distributed and Symbolic Execution for Natural Language Queries}

\newcommand{\newcite}[1]{\citeauthor{#1}~(\citeyear{#1})}

\DeclareMathOperator{\softmax}{softmax}
\DeclareMathOperator{\MLP}{MLP}
\DeclareMathOperator{\sigmoid}{sigmoid}

\begin{document} 

\twocolumn[
\icmltitle{Coupling Distributed and Symbolic Execution for Natural Language Queries}



\icmlsetsymbol{equal}{*}

\begin{icmlauthorlist}
	\icmlauthor{Lili Mou}{pku}
	\icmlauthor{Zhengdong Lu}{deeply}
	\icmlauthor{Hang Li}{huawei}
	\icmlauthor{Zhi Jin}{pku}
\end{icmlauthorlist}

\icmlaffiliation{pku}{Key Laboratory of High Confidence Software Technologies (Peking University),
	MoE; Software Institute, Peking University, China}
\icmlaffiliation{deeply}{DeeplyCurious.ai}
\icmlaffiliation{huawei}{Noah's Ark Lab, Huawei Technologies.\quad Work done when the first author was an intern at Huawei}

\icmlcorrespondingauthor{L.M.}{double power.mou@gmail.com}
\icmlcorrespondingauthor{Z.L.}{luz@DeeplyCurious.ai}
\icmlcorrespondingauthor{H.L.}{HangLi.HL@huawei.com}
\icmlcorrespondingauthor{Z.J.}{zhijin@sei.pku.edu.cn}

\icmlkeywords{semantic parsing, neural execution, symbolic execution}

\vskip 0.3in
]



\printAffiliationsAndNotice{}  

\begin{abstract} 
Building neural networks to query a knowledge base (a table) with natural language is an emerging research topic in deep learning. An executor for table querying typically requires multiple steps of execution because queries may have complicated structures. In previous studies, researchers have developed either fully distributed executors or symbolic executors for table querying. A distributed executor can be trained in an end-to-end fashion, but is weak in terms of execution efficiency and explicit interpretability. A symbolic executor is efficient in execution, but is very difficult to train especially at initial stages. 
In this paper, we propose to couple distributed and symbolic execution for natural language queries, where the symbolic executor is pretrained with the distributed executor's intermediate execution results in a step-by-step fashion. Experiments show that our approach significantly outperforms both distributed and symbolic executors, exhibiting high accuracy, high learning efficiency, high execution efficiency, and high interpretability.
\end{abstract}

\section{Introduction}

Using natural language to query a knowledge base is an important task in NLP and has wide applications in question answering (QA)~\cite{neuralQA}, human-computer conversation~\cite{dialog}, etc. Table~\ref{tab:example} illustrates an example of a knowledge base (a table) and a query ``How long is the game with the largest host country size?'' To answer the question, we should first find a row with the largest value in the column \textit{Area}, and then select the value of the chosen row with the column being \textit{Duration}.

A typical approach to table querying is to convert a natural language sentence to an ``executable'' logic form, known as \textit{semantic parsing}. Traditionally, building a semantic parser requires extensive human engineering of explicit features~\cite{FB,sempre}.

With the fast development of deep learning, an increasingly number of studies use neural networks for semantic parsing. 
\newcite{NN} and \newcite{NN2} apply sequence-to-sequence (\texttt{seq2seq}) neural models to generate a logic form conditioned on an input sentence, but the training requires  groundtruth logic forms, which are costly to obtain and specific to a certain dataset. In realistic settings, we only assume groundtruth denotations\footnote{A \textit{denotation} refers to an execution result.} are available, and that we do not know execution sequences or intermediate execution results.
\newcite{symbolic} train a \texttt{seq2seq} network by REINFORCE policy gradient. But it is known that the REINFORCE algorithm is sensitive to the initial policy; also, it could be very difficult to get started at early stages.

\begin{table}[!t]
	\small
	\textbf{Query}:\\
	\verb|  | How long is the game with the largest host country size?\\[-.3cm]

	\textbf{Knowledge base (table)}:
	\begin{center}
		\resizebox{.8\linewidth}{!}{
			\begin{tabular}{|c|c|c|c|c|c|}
				\hline
				\textbf{Year} & \textbf{City} & $\cdots$ & \textbf{Area} &$\cdots$& \textbf{Duration}\\
				\hline\hline
				\multicolumn{6}{|c|}{$\cdots$}  \\
				\hline
				2000  &      Sydney    &   $\cdots$ & 200         & $\cdots$    &    30\\
				\hline 2004  &      Athens    &   $\cdots$ & 250         & $\cdots$    &    20\\
				\hline
				2008  &       Beijing   &  $\cdots$ & 350         & $\cdots$    &   25\\
				\hline
				2012  &       London    &  $\cdots$ & 300         & $\cdots$    &   35\\      
				\hline       
				2016  & Rio de Janeiro  &  $\cdots$ & 200         & $\cdots$    &   40   \\
				\hline
				\multicolumn{6}{|c|}{$\cdots$}  \\
				\hline
			\end{tabular}
		}
	\end{center}
	\vspace{-.2cm}
	\caption{An example of a natural language query and a knowledge base (table).}\label{tab:example}
	\vspace{-.4cm}
\end{table}

\newcite{NE} propose a fully distributed neural enquirer, comprising several neuralized execution layers of field attention, row annotation, etc. The model can be trained in an end-to-end fashion because all components are differentiable. However, it lacks explicit interpretation and is not efficient in execution due to intensive matrix/vector operation during neural processing.

\newcite{programmer} propose a neural programmer by defining a set of symbolic operators (e.g., \texttt{argmax}, \texttt{greater\_than}); at each step, all possible execution results are fused by a softmax layer, which predicts the probability of each operator at the current step. The step-by-step fusion is accomplished by weighted average and the model is trained with mean square error. Hence, such approaches work with numeric tables, but may not be suited for other operations like string matching. It also suffers from the problem of ``exponential numbers of combinatorial states,'' as the model explores the entire space at a time by step-by-step weighted average. 

In this paper, we propose to couple distributed and symbolic execution for natural language queries. By ``symbolic execution,'' we mean that  we define symbolic operators and keep discrete intermediate execution results.\footnote{The sense of \textit{symbolic execution} here should not be confused with ``symbolic execution of a program'' (see \url{https://en.wikipedia.org/wiki/Symbolic_execution} for example).} Our intuition rises from the observation that a fully distributed/neuralized executor also exhibits some (imperfect) symbolic interpretation. For example, the field attention gadget in \newcite{NE} generally aligns with column selection.
We therefore use the distributed model's intermediate execution results as supervision signals to pretrain a symbolic executor. Guided by such imperfect step-by-step supervision, the symbolic executor learns a fairly meaningful initial policy, which largely alleviates the cold start problem of REINFORCE.
Moreover, the improved policy can be fed back to the distributed executor to improve the neural network's performance.

We evaluated the proposed approach on the QA dataset in~\newcite{NE}. Our experimental results show that the REINFORCE algorithm alone takes long to get started. Even if it does, it is stuck in poor local optima. Once pretrained by imperfect supervision signals, the symbolic executor can recover most execution sequences, and also achieves the highest denotation accuracy.
It should be emphasized that, in our experiment, neither the distributed executor nor the symbolic executor is aware of groundtruth execution sequences, and that the entire model is trained with weak supervision of denotations only. 

To the best of our knowledge, we are the first to couple distributed and symbolic execution for semantic parsing. Our study also sheds light on neural sequence prediction in general.

\begin{figure}[!t]
	\centering
	\includegraphics[width=\linewidth]{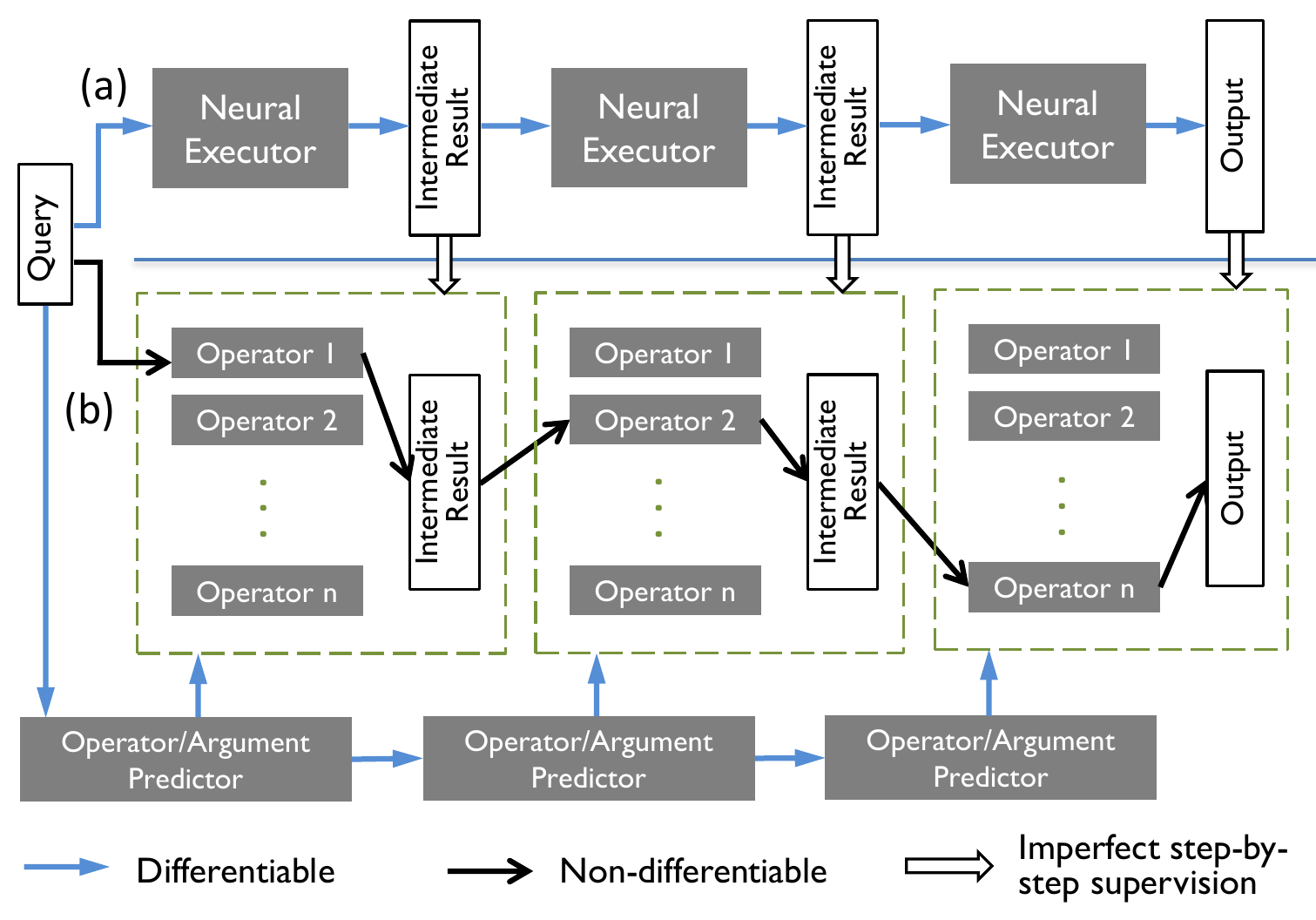}\vspace{-.3cm}
	\caption{An overview of the coupled distributed and symbolic executors.}\label{fig:couple}
	\vspace{-.3cm}
\end{figure}

\section{Approach}
In Subsection~\ref{ss:distributed}, we introduce the fully distributed neural executor, which is mostly based on \newcite{NE}.
For symbolic execution, we design a set of operators that are complete to the task at hand; at each execution step, a neural network predicts a particular operator and possibly arguments (Subsection~\ref{ss:symbolic}).

Subsection~\ref{ss:unified} provides a unified view of distributed and symbolic execution (Figure~\ref{fig:couple}). We explain how the symbolic executor is pretrained by the distributed one's intermediate execution results, and then further trained with the REINFORCE algorithm.

\subsection{Distributed Executor}\label{ss:distributed}

The distributed executor makes full use of neural networks for table querying.
By ``distributed,'' we mean that all semantic units (including words in the query, entries in the table, and execution results) are represented as distributed, real-valued vectors and processed by neural networks.
One of the most notable studies of distributed semantics is word embeddings, which map discrete words to vectors as meaning representations~\cite{wordembed}.

\begin{figure}[!t]
	\centering
	\includegraphics[width=.75\linewidth]{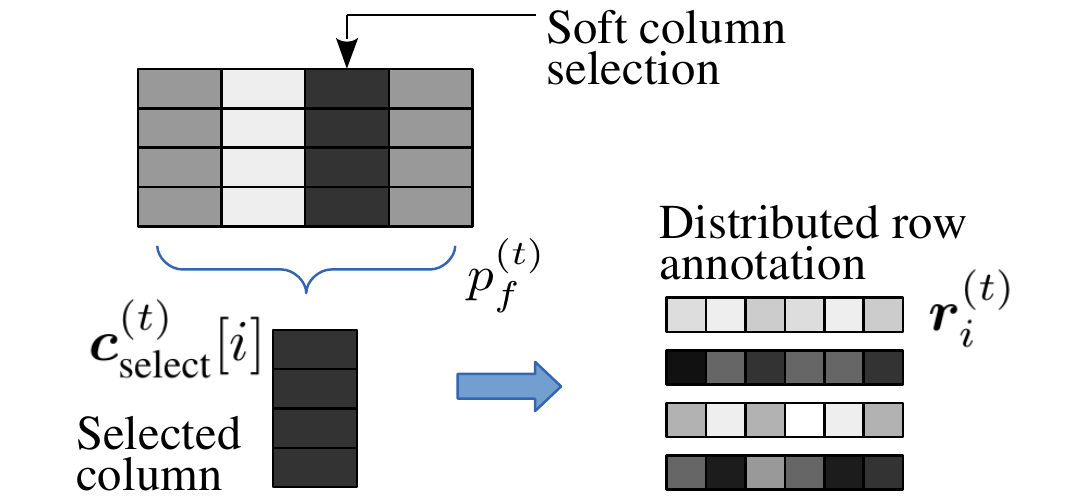}
	\vspace{-1mm}
	\caption{A single step of distributed execution.}\label{fig:distributed}
	\vspace{-2mm}
\end{figure}

The distributed executor consists of the following main components.
\begin{compactitem}
	\item Query encoder. Words are mapped to word embeddings and a bi-directional recurrent neural network (RNN) aggregates information over the sentence. RNNs' last states in both directions are concatenated as the query representation (detonated as $\bm q$).
	\item Table encoder. All table cells are also represented as embeddings. For a cell $c$ (e.g., \textit{Beijing} in Table~\ref{tab:example}) with its column/field name being $f$ (e.g., \textit{City}), the cell vector is the concatenation of the embeddings of $c$ and $f$, further processed by a feed-forward neural network (also known as multi-layer perceptron). We denote the representation of a cell as $\bm c$.
	\item Executor. As shown in Figure~\ref{fig:couple}a, the neural network comprises several steps of execution. In each execution step, the neural network selects a column by softmax attention, and annotates each row with a vector, i.e., an embedding (Figure~\ref{fig:distributed}). The row vector can be intuitively thought of as row selection in query execution, but is represented by distributed semantics here.\footnote{In a pilot experiment, we tried a gating mechanism to indicate the results of row selection in hopes of aligning symbolic table execution. However, our preliminary experiments show that such gates do not exhibit much interpretation, but results in performance degradation. The distributed semantics provide more information than a 1-bit gate for a row.
	} 
	In the last step of execution, a softmax classifier is applied to the entire table to select a cell as the answer. Details are further explained as below.
\end{compactitem}

\begin{table*}[!t]
	\centering
	\resizebox{.96\linewidth}{!}{
		\begin{tabular}{|ll|}
			\hline
			\textbf{Operator} & \textbf{Explanation}       \\
			\hline\hline
			\texttt{select\_row} & Choose a row whose value of a particular column is mentioned in the query \\
			\texttt{argmin}      & Choose the row from previously selected candidate rows 
			with the minimum value in a particular column\\
			\texttt{argmax}      & Choose the row from previously selected candidate rows 
			with the maximum value in a particular column\\         
			\texttt{greater\_than} & Choose rows whose value in a particular column is greater than a previously selected row\\  
			\texttt{less\_than} & Choose rows whose value in a particular column is less than a previously selected row\\ 
			\texttt{select\_value} & Choose the value of a particular column  and of the previously selected row\\    
			\texttt{EOE} & Terminate, indicating the end of execution\\
			\hline
		\end{tabular}
	}\vspace{-2mm}
	\caption{Primitive operators for symbolic execution.}\label{tab:symbolic}
	\vspace{-2mm}
\end{table*}

Let $\bm r_i^{(t-1)}$ be the previous step's row annotation result, where the subscript $i$ indexes a particular row. 
We summarize  global execution information (denoted as $\bm g^{(t-1)}$) by max-pooling the row annotation $\bm r^{(t-1)}$, i.e.,
\begin{equation}
\bm g^{(t-1)} = \operatorname{MaxPool}_i \left\{\bm r_i^{(t-1)}\right\}
\end{equation}
In the current execution step, we first compute a distribution $p_{f}^{(t)}$ over all fields as ``soft'' field selction. The computation is based on the query~$\bm q$, the previous global information $\bm g^{(t-1)}$, and the field name embeddings~$\bm f$, i.e.,
\begin{align}
p_{f_j}^{(t)}&=\softmax\left(\MLP\big([\bm q; \bm f_j; \bm g^{(t-1)}]\big)\right)\\
&=\frac{\exp\left\{
	\MLP\left([\bm q; \bm f_j; \bm g^{(t-1)}]\right)
	\right\}
}{   \sum_{j'} \exp\left\{
\MLP\left([\bm q; \bm f_{j'}; \bm g^{(t-1)}]\right)
\right\}
}
\label{eqn:fieldselection}
\end{align}
where $[\cdot;\ \cdot;\ \cdots]$ denotes vector concatenation; MLP refers to a multi-layer perceptron.

Here, the weights of softmax are field embeddings (rather than parameters indexed by positions). In this way, there is no difference if one shuffles table columns. Besides, for a same field name, its embedding is shared among all training samples containing this field (but different tables may have different fields).

We represent the selected cell in each row as the sum of all cells in that row, weighted by soft field selection $p_f^{(t-1)}$. Formally, for the $i$-th row, we have
\begin{equation}
\bm c_\text{select}^{(t)}[i]=\sum_j p_{f_j}^{(t)}\bm c_{ij}
\end{equation}
where $[i]$ indexes a row of the selected column.

The current row annotation is computed by another MLP, based on the query~$\bm q$, previous execution results~$\bm r_i^{(t-1)}$, previous global information $\bm g^{(t-1)}$, as well as the selected row in the current step $\bm c_\text{select}^{(t)}[i]$ (the selection is in a soft manner), i.e., 
\begin{equation}
\bm r_i^{(t)}=\MLP\left(\left[\bm q, \bm g^{(t-1)}, \bm r_i^{(t-1)}, \bm c_\text{select}^{(t)}[i]\right]\right)
\end{equation}

As said, the last execution layer applies a softmax classifier over all cells to select an answer.
Similar to Equation~\ref{eqn:fieldselection}, the weights of softmax are not associated with positions, but the cell embeddings. In other words, the probability of choosing the $i$-th row, $j$-th column is
\begin{equation}
p_{ij} =\frac{\exp\left\{\MLP\left(\bm q, \bm g^{(t-1)}, \bm r_i^{(t-1)}, \bm c_{ij}\right)\right\}}
{\sum_{i'}\sum_{j'}\exp\left\{
	\MLP\left(\bm q, \bm g^{(t-1)}, \bm r_{i'}^{(t-1)}, \bm c_{i'j'}\right)
	\right\}}
\end{equation}

In this way, the neural executor is invariant with respect to the order of rows and columns.
While order-sensitive architectures (e.g., convolutional/recurrent neural networks) might also model a table by implicitly ignoring such order information, the current treatment is more reasonable, which also better aligns with symbolic interpretation.

\subsection{Symbolic Executor}\label{ss:symbolic}

The methodology of designing a symbolic executor is to define a set of primitive operators for the task, and then to use a machine learning model to predict the operator sequence and its arguments.

Our symbolic executor is different from the neural programmer \cite{programmer} in that we keep discrete/symbolic operators as well as execution results, whereas \newcite{programmer} fuse execution results by weighted average.
\subsubsection{Primitive Operators}\label{ss:operator}
We design six operators for symbolic execution, which are complete as they cover all types of queries in our scenario. Similar to the distributed executor, the result of one-step symbolic execution is some information for a row; here, we use a 0-1 boolean scalar, indicating whether a row is selected or not after a particular step of execution. Then a symbolic execution step takes previous results as input, with a column/field being the argument. The green boxes in Figure~\ref{fig:couple}b illustrate the process and Table~\ref{tab:symbolic} summarizes our primitive operator set.

In Table~\ref{tab:example}, for example, the first step of execution is \texttt{argmax} over the column \textit{Area}, with previous (initial) row selection being all ones. This step yields a single row $\langle 2008, \text{Beijing}, \cdots, 350, \cdots, 25\rangle$. The second execution operator is \texttt{select\_value}, with an argument column=\texttt{Duration}, yielding the result 25. Then the executor terminates (\texttt{EOE}).

Stacked with multiple steps of primitive operators, the executor can answer fairly complicated questions like ``How long is the last game which has smaller country size than the game whose host country GDP is 250?'' In this example, the execution sequence is
\begin{compactenum}
	\item \texttt{select\_row}: select the row where the column is \textit{GDP} and the value is mentioned in the query. 
	\item \texttt{less\_than}: select rows whose country size is less than that of the previously selected row. 
	\item \texttt{argmax}: select the row whose year is the largest among previously selected rows.
	\item \texttt{select\_value}: choose the value of the previously selected row with the column being \textit{Duration}.
\end{compactenum}
Then the execution terminates. In our scenario, the execution is limited to four steps (\texttt{EOE} excluded) as such queries are already very complicated in terms of logical depth.

\subsubsection{Operator and Argument Predictors}\label{ss:predictor}

We also leverage neural models, in particular RNNs, to predict the operator and its argument (a selected field/column). 

Let $\bm h^{(t-1)}$ be the previous state's hidden vectors. The current hidden state is 
\begin{equation}
\bm h_\text{op}^{(t)} = \sigmoid(W_\text{op}^\text{(rec)}\bm h_\text{op}^{(t-1)})
\end{equation}
\noindent where $W_\text{op}^\text{(rec)}$ is weight parameters. (A bias term is omitted in the equation for simplicity.) The initial hidden state is the query embedding, i.e., $\bm h_\text{op}^{(0)}=\bm q$.

The predicted probability of an operator $i$ is given by
\begin{equation}
p_{\text{op}_i}^{(t)} = \softmax\left\{\bm w_{\text{op}_i}^\text{(out)}{}^\top\bm h_\text{op}^{(t)}\right\}\label{eqn:op_softmax}
\end{equation}
The operator with the largest predicted probability is selected for execution.

Our RNN here does not have input, because the execution sequence is not dependent on the result of the previous execution step. Such architecture is known as a Jordan-type RNN~\cite{jordan,jordan2}.

Likewise, another Jordan-type RNN  selects a field. The only difference lies in
the weight of the output softmax, i.e., $\bm w_i$ in Equation~\ref{eqn:op_softmax} is substituted with the  embedding of a field/column name $\bm f$, given by
\begin{align}
\bm h_\text{field}^{(t)} &= \sigmoid(W_\text{field}^\text{(rec)}\bm h_\text{field}^{(t-1)})\\
p_{f_j}^{(t)} &= \softmax\left\{\bm  f_j^\top\bm h_\text{field}^{(t)}\right\}\label{eqn:fd_softmax}
\end{align}

Training a symbolic executor without step-by-step supervision signals is non-trivial.
A typical training method is reinforcement learning in a trial-and-error fashion. However, for a random initial policy, the probability of recovering an accurate execution sequence is extremely low. Given a $10\times10$ table, for example, the probability is $1/(6^4\cdot 10^4)\approx7.7\times10^{-8}$; the probability of obtaining an accurate denotation is 1\%, which is also very low. Therefore, symbolic executors are not efficient in learning.

\subsection{A Unified View}\label{ss:unified}

We now have two worlds of execution: 
\begin{compactitem}
	\item The distributed executor is end-to-end learnable, but it is of low execution efficiency  because of intensive matrix/vector multiplication during neural information processing. The fully neuralized execution also lacks explicit interpretation.
	\item The symbolic executor has high execution efficiency and explicit interpretation. However, it  cannot be trained in an end-to-end manner, and suffers from the cold start problem of reinforcement learning.
\end{compactitem}

We propose to combine the two worlds by using the distributed executor's intermediate execution results to pretrain the symbolic executor for an initial policy; we then use the REINFORCE algorithm to improve the policy. 
The well-trained symbolic executor's intermediate results can also be fed back to the distributed executor to improve performance.
\subsubsection{Distributed $\rightarrow$ Symbolic}\label{ss:DS}

We observe that the field attention in Equation~\ref{eqn:fieldselection} generally aligns with column selection in Equation~\ref{eqn:fd_softmax}. We therefore pretrain the column selector in the symbolic executor with labels predicted by a fully neuralized/distributed executor.
Such pretraining can obtain up to 70\% accurate field selection and largely reduce the search space during reinforcement learning. 

Formally, the operator predictor (Equation~\ref{eqn:op_softmax}) and argument predictor (Equation~\ref{eqn:fd_softmax}) in each execution step are the \textit{actions} (denoted as $\mathcal{A}$) in reinforcement learning terminologies. If we would like to pretrain $m$ actions $a_1, a_2,\cdots, a_m\in\mathcal{A}$, the cost function of a particular data sample is
\begin{equation}
J=-\sum_{i=1}^m\sum_{j=1}^{n_{\text{label}}^{(i)}} \hat{t}_j^{(i)} \log p_j^{(i)}\label{eqn:fieldloss}
\end{equation}
where $n_{\text{label}}^{(j)}$ is the number of labels (possible candidates) for the $j$-th action.
$\bm p^{(i)}\in\mathbb{R}^{n_{\text{label}}^{(i)}}$ is the predicted probability by the operator/argument predictors in Figure~\ref{fig:couple}b. 
$\hat{\bm t}^{(i)}\in\mathbb{R}^{n_{\text{label}}^{(i)}}$ is the induced action from the fully distributed model in Figure~\ref{fig:couple}a. In our scenario, we only pretrain column predictors.

After obtaining a meaningful, albeit imperfect, initial policy, we apply REINFORCE~\cite{REINFORCE} to improve the policy. 
\begin{table*}[!t]
	\centering
	\resizebox{.93\linewidth}{!}{
		\begin{tabular}{|l|c|ccc|ccc|}
			\hline
			&              &\multicolumn{3}{|c|}{\textbf{Denotation}} &\multicolumn{3}{|c|}{\textbf{Execution}}\\
			\textbf{Query type} & \sc Sempre \hspace{-1.4cm} Sempre$^\dag$ & \textbf{Distributed}$^\dag$ & \textbf{Symbolic}& \textbf{Coupled} & \textbf{Distributed} & \textbf{Symbolic} &\textbf{Coupled}\\
			\hline\hline
			\tt SelectWhere &     93.8  & 96.2   & \ \ 99.2       & \textbf{\ \ 99.6} & -- &\ \ 99.1&\textbf{\ \ 99.6} \\
			\tt Superlative &     97.8  & {98.9} & \textbf{100.0}       & \textbf{100.0}& -- & \textbf{100.0} & \textbf{100.0}\\
			\tt WhereSuperlative & 34.8 & 80.4   & 51.9       &  \textbf{\ \ 99.9} & --& \ \ \ \ 0.0 & \textbf{\ \ 91.0} \\
			\tt NestQuery     &    34.4 & 60.5  & 52.5 &  \textbf{100.0}   &  -- &\ \ \ \ 0.0&\textbf{100.0}\\
			\hline
			Overall      &     65.2 & 84.0  & 75.8 & \textbf{\ \ 99.8} & -- &\ \ 49.5&\textbf{\ \ 97.6}\\
			\hline
		\end{tabular}}\vspace{-1.5mm}
		\caption{Accuracies (in percentage) of the \textsc{Sempre} tookit, the distributed neural enquirer, the symbolic executor, and our coupled approach. \protect$^\dag$Results reported in~\newcite{NE}.}\label{tab:result}
		\vspace{-2.5mm}
	\end{table*}
	
	We define a binary reward $R$ indicating whether the final result of symbolic execution matches the groundtruth denotation. The loss function of a policy is the negative expected reward where actions are sampled from the current predicted probabilities
	\begin{equation}
	J=-\mathbb{E}_{a_1, a_2, \cdots, a_n\sim \theta}[R(a_1, a_2, \cdots, a_n)]
	\end{equation}
	The partial derivative for a particular sampled action is 
	\begin{equation}\label{eqn:derivative}
	\dfrac{\partial J}{\partial\bm o_i}=\tilde{R}\cdot (\bm p_i - \mathbf{1}_{a_i})
	\end{equation}
	where $\bm p_i$ is the predicted probability of all possible actions at the time step $i$, $\mathbf{1}_{a_i}$ is a onehot representation of a sampled action $a_i$, and $\bm o_i$ is the input (also known as \textit{logit}) of softmax.
	$\tilde{R}$ is the adjusted reward, which will be described shortly.
	
	To help the training of REINFORCE, we have two tricks:
	\begin{compactitem}
		\item We balance exploration and exploitation with a small probability $\epsilon$. In other words, we sample an action from the predicted action distribution with probability $1-\epsilon$ and from a uniform distribution over all possible actions with probability $\epsilon$. 
		The small fraction of uniform sampling helps the model to escape from poor local optima, as it continues to explore the entire action space during training.
		\item We adjust the reward by subtracting the mean reward, averaged over sampled actions for a certain data point. This is a common practice for REINFORCE~\cite{sequencelevel}. We also truncate negative rewards as zero to prevent gradient from being messed up by incorrect execution. This follows the idea of ``reward-inaction,'' where unsuccessful trials are ignored (Section~2.4 in \newcite{RL}). The adjusted reward is denoted as $\tilde R$ in Equation~\ref{eqn:derivative}.
	\end{compactitem}
	Notice that these tricks are applied to both coupled training and baselines for fairness.
	\subsubsection{Distributed $\rightarrow$ Symbolic $\rightarrow$ Distributed}\label{ss:DSD}
	After policy improvement by REINFORCE, we could further feed back the symbolic executor's intermediate results to the distributed one, akin to the step-by-step supervision setting in \newcite{NE}. The loss is a combination of denotation cross entropy loss $J_\text{denotation}$ and field attention cross entropy loss $J_\text{fields}$. ($J_\text{fields}$ is similar to Equation~\ref{eqn:fieldloss}, and details are not repeated). The overall training objective is $J=J_\text{denotation} +\lambda J_\text{fields}$, where $\lambda$ is a hyperparameter balancing the two factors.
	
	As will be seen in Section~\ref{ss:feedback}, feeding back intermediate results improves the distributed executor's performance. This shows the distributed and symbolic worlds can indeed be coupled well.
	\section{Experiments}

	\subsection{Dataset}
	We evaluated our approach on a QA dataset in~\newcite{NE}. The dataset comprises 25k different tables and queries for training; validation and test sets contain 10k samples, respectively, and do not overlap with the training data. Each table is of size $10\times 10$, but different samples have different tables; the queries can be divided into four types: \texttt{SelectWhere}, \texttt{Superlative}, \texttt{WhereSuperlative}, and \texttt{NestQuery}, requiring 2--4 execution steps (\texttt{EOE} excluded). 
	
	We have both groundtruth denotation and execution actions (including operators and fields), as the dataset is synthesized by complicated rules and templates for research purposes.  However, only denotations are used as labels during training, which is a realistic setting; execution sequences are only used during testing.
	For the sake of simplicity, we presume the number of execution steps is known \textit{a priori} during training (but not during testing). Although we have such (little) knowledge of execution, it is not a limitation of our approach and out of our current focus. One can easily design a \texttt{dummy} operator to fill an unnecessary step or one can also train a discriminative sentence model to predict the number of execution steps if a small number of labels are available.
	
	We chose the synthetic datasets because it is magnitudes larger than existing resources (e.g., \textsc{WebQuestions}). The process of data synthesizing also provides intermediate execution results for in-depth analysis. Like \textsc{bAbI} for machine comprehension, our dataset and setting are a prerequisite for general semantic parsing.
	The data are available at out project website\footnote{\url{https://sites.google.com/site/coupleneuralsymbolic/}};
	the code for data generation can also be downloaded to facilitate further development of the dataset.
	
	\subsection{Settings}
	The symbolic executor's settings were generally derived from~\newcite{NE} so that we can have a fair comparison.\footnote{One exception is the query RNN's hidden states. \newcite{NE} used 300d BiRNN, but we found it more likely to overfit in the symbolic setting, and hence we used 50d. This results in slower training, more rugged error surfaces, but higher peak performance.} The dimensions of all layers were in the range of 20--50; the learning algorithm was AdaDelta with default hyperparameters.
	
	For the pretraining of the symbolic executor, we applied maximum likelihood estimation for 40 epochs to column selection with labels predicted by the distributed executor. We then used the REINFORCE algorithm to improve the policy, where we generated 10 action samples for each data point with the exploration probability $\epsilon$ being $0.1$.
	
	When feeding back to the distributed model, we chose $\lambda$ from $\{0.1, 0.5, 1\}$ by validation to balance denotation error and field attention error. $0.5$ outperforms the rest.
	
	Besides neural networks, we also included the \textsc{Sempre} system as a baseline for comparison. The results are reported in \newcite{NE}, where a \textsc{Sempre} version that is specially optimized for table query is adopted~\cite{sempre}. Thus it is suited in our scenario.
	\subsection{Results}
	
	\subsubsection{Overall Performance}
	Table~\ref{tab:result} presents the experimental results of our coupled approach as well as baselines.
	Because reinforcement learning is much more noisy to train, we report the test accuracy corresponding to highest validation accuracy among three different random initializations (called \textit{trajectories}). This is also known as a \textit{restart} strategy for non-convex optimization.
	
	As we see, both distributed and symbolic executors outperform the \textsc{Sempre} system, showing that neural networks can capture query information more effectively than human-engineered features.
	Further, the coupled approach also significantly outperforms both of them.
	
	If trained solely by REINFORCE, the symbolic executor can recover the execution sequences for simple questions (\texttt{SelectWhere} and \texttt{Superlative}). However, for more complicated queries, it only learns last one or two steps of execution and has trouble in recovering early steps, even with the tricks in Section~\ref{ss:DS}. This results in low execution accuracy but near 50\% denotation accuracy because, in our scenario, we still have half chance to obtain an accurate denotation even if the nested (early) execution is wrong---the ultimate result is either in the candidate list or not, given a wrong where-clause execution.
	
	By contrast, the coupled training largely improves the symbolic executor's performance in terms of all query types.
	
	\subsubsection{Interpretability}
	The accuracy of execution is crucial to the interpretability of a model. We say an execution is \textit{accurate}, if all actions (operators and arguments) are correct.
	As shown above, an accurate denotation does not necessarily imply an accurate execution. 
	
	We find that coupled training recovers most correct execution sequences, that the symbolic executor alone cannot recover complicated cases, and that a fully distributed enquirer does not have explicit interpretations of execution.
	
	The results demonstrate high interpretability of our approach, which is helpful for human understanding of execution processes.
	
	\subsubsection{Learning Efficiency}
\begin{figure}[!t]
		\centering
		\includegraphics[width=.45\linewidth]{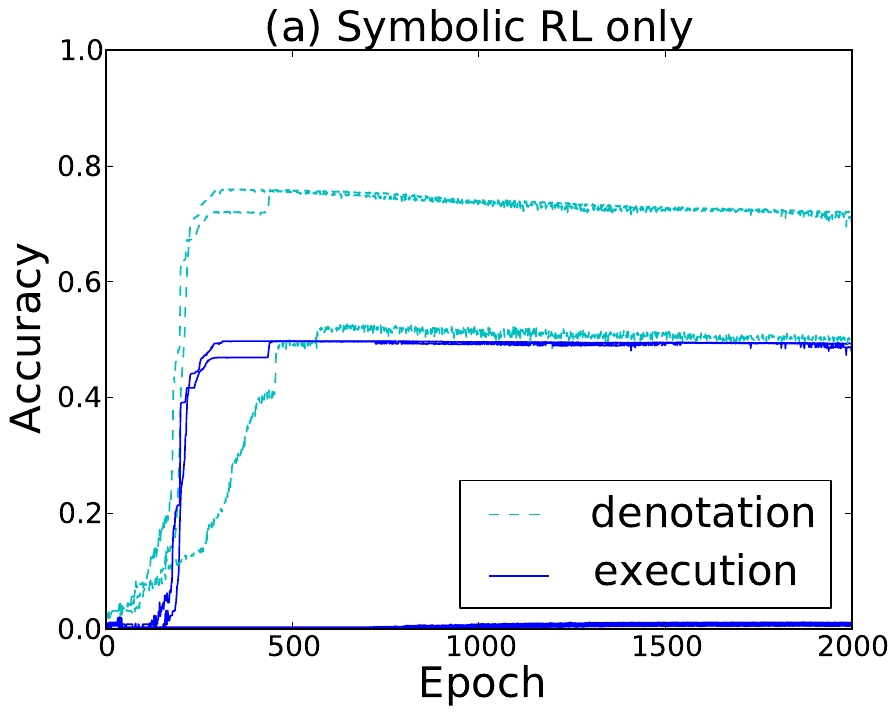} \includegraphics[width=.45\linewidth]{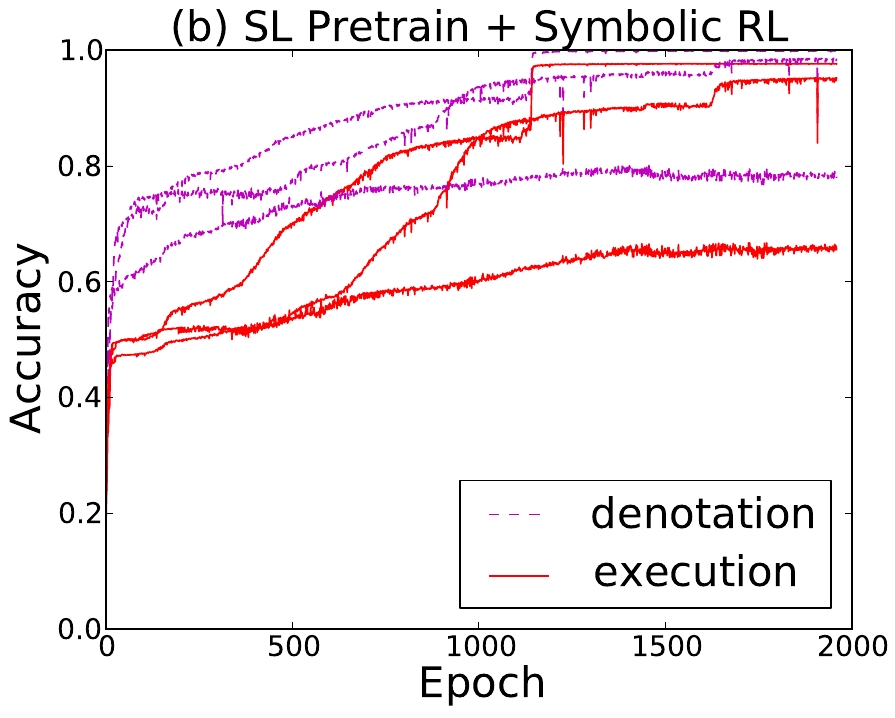}
		\vspace{-.25cm}
		\caption{Validation learning curves. (a) Symbolic executor trained by REINFORCE (RL) only. (b) Symbolic executor with 40-epoch pretraining using a distributed executor in a supervised learning (SL) fashion.  Both settings have three trajectories with different random initializations. Dotted lines: Denotation accuracy. Solid lines: Execution accuracy.}
		\label{fig:curve}
		\vspace{-.25cm}
	\end{figure}

	We plot in Figure~\ref{fig:curve} the validation learning curves of the symbolic executor, trained by either reinforcement learning alone or our coupled approach.
	
	Figure~\ref{fig:curve}a shows that the symbolic executor is hard to train by REINFORCE alone: one trajectory obtains near-zero execution accuracy after 2000 epochs; the other two take 200 epochs to get started to escape from initial plateaus. Even if they achieve $\sim$50\% execution accuracy (for simple query types) and $\sim$75\% denotation accuracy, they are stuck in the poor local optima.
	
	Figure~\ref{fig:curve}b presents the learning curves of the symbolic executor pretrained with intermediate field attention of the distributed executor. Since the operation predictors are still hanging after pretraining, the denotation accuracy is near 0 before reinforcement learning. However, after only a few epochs of REINFORCE training, the performance increases sharply and achieves high accuracy gradually.
	
	Notice that we have 40 epochs of (imperfectly) supervised pretraining. However, its time is negligible compared with reinforcement learning because in our experiments REINFORCE generates 10 samples and hence is theoretically 10 times slower.   
	The results show that our coupled approach has much higher learning efficiency than a pure symbolic executor.
	
	\subsubsection{Execution Efficiency}

	Table~\ref{tab:efficiency} compares the execution efficiency of a distributed executor and our coupled approach. All neural networks are implemented in Theano with  a TITAN Black GPU and Xeon e7-4820v2 (8-core) CPU; symbolic execution is assessed in \texttt{C++} implementation. The comparison makes sense because the Theano platform is not specialized in symbolic execution, and fortunately, execution results do not affect actions in our experiment. Hence they can be easily disentangled.
	
	As shown in the table, the execution efficiency of our approach is 2--5 times higher than the distributed executor, depending on the implementation. The distributed executor is when predicting because it maps every token to a distributed real-valued vector, resulting in intensive matrix-vector operations.
	The symbolic executor only needs a neural network to predict actions (operators and arguments), and thus is more lightweight. Further, we observe the execution itself is blazingly fast, implying that, compared with distributed models, our approach could achieve even more efficiency boost with a larger table or more complicated operation.
	
	\begin{table}[!t]
		\resizebox{\linewidth}{!}{
			\begin{tabular}{|c||c||c|c|c|}
				\hline
				&  \textbf{Fully} & \multicolumn{3}{c|}{\textbf{Our approach}}\\
				\cline{3-5}
				&  \textbf{Distributed}              &\!\!\!\! Op/Arg Pred. \!\!\!\!&\!\!\!\! Symbolic Exe.$^\dag$ \!\!\!\!&  Total\\ 
				\hline\hline
				\!\!\textbf{CPU}\!\! &  13.86 & 2.65 & \multirow{2}{*}{0.002}& 2.65\\
				\cline{1-3}\cline{5-5}
				\!\!\textbf{GPU}\!\! &  \ \ 1.05 & 0.44 &  & 0.44\\
				\hline
			\end{tabular}
		}
		\caption{Execution efficiency. We present the running time (in seconds) of the test set, containing 10k samples. $^\dag$The symbolic execution is assessed in \texttt{C++} implementation. Others are implemented in Theano, including the fully distributed model as well as the operator and argument predictors.}\label{tab:efficiency}
	\end{table}
	
	\begin{table}
		\centering
		\resizebox{!}{!}{
			\begin{tabular}{lr}
				\toprule
				Training Method & Accuracy ($\%$)\\
				\midrule
				End-to-end  (w/ denotation labels)$^\dag$   &    84.0\ \ \\ 
				Step-by-step (w/ execution labels)$^\dag$   &    96.4\ \ \\
				Feeding back                         &    96.5\ \ \\
				\bottomrule
			\end{tabular}
		}
		\caption{The accuracy of a fully distributed model, trained by different methods. In the last row, we first train a distributed executor and feed its intermediate execution results to the symbolic one; then the symbolic executor's intermediate results are fed back to the distributed one. $^\dag$Reported in \protect\newcite{NE}.}\label{tab:feedback}
	\end{table}
	
	\begin{figure*}
		\centering
		\textbf{Query}: How many people watched the earliest game whose host country GDP is larger than the game in Cape Town?
		\includegraphics[width=.95\linewidth]{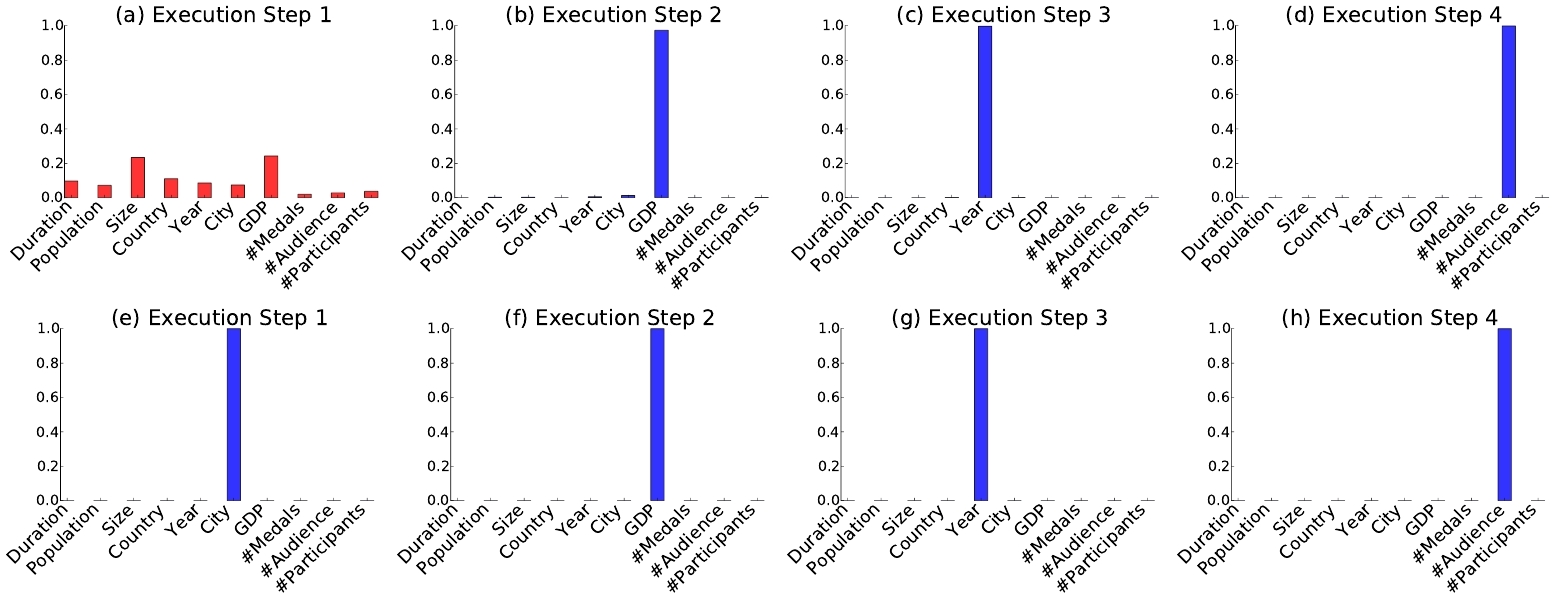}
		\vspace{-2mm}
		\caption{Distributed executor's intermediate results of field attention. \textbf{Top}: Trained in an end-to-end fashion (a--d). \textbf{Bottom}: One-round co-training of distributed and symbolic executors (e--h). The {\color{red} red} plot indicates incorrect field attention.}
		\vspace{-2mm}
		\label{fig:showcase}
	\end{figure*}
	
	\subsubsection{Feeding back}\label{ss:feedback}
	
	We now feed back the well-trained symbolic executor's intermediate results to the fully distributed one. As our well-trained symbolic executor has achieved high execution accuracy,
	this setting is analogous to strong supervision with step-by-step groundtruth signals, and thus it also achieves similar performance,\footnote{We even have 0.1\% performance boost compared with the step-by-step setting, but we think it should be better explained as variance of execution.} shown in Table~\ref{tab:feedback}.
	
	We showcase the distributed executor's field attention\footnote{The last neural executor is a softmax layer over all cells. We marginalize over rows to obtain the field probability.} in Figure~\ref{fig:showcase}. If trained in an end-to-end fashion, the neural network exhibits interpretation in the last three steps of this example, but in the early step, the field attention is incorrect (also more uncertain as it scatters a broader range). After feeding back the symbolic executor's intermediate results as step-by-step supervision, the distributed executor exhibits near-perfect field attention.
	
	This experiment further confirms that the distributed and symbolic worlds can indeed be coupled well. In more complicated applications, there could also be possibilities in iteratively training one model by leveraging the other in a co-training fashion.

	\section{Related Work and Discussions}
	
	Neural execution has recently aroused much interest in the deep learning community.
	Besides SQL-like execution as has been extensively discussed in previous sections,
	neural Turing machines \cite{NTM} and neural programmer-interpreters~\cite{NPI} are aimed at more general ``programs.'' 
	The former is a ``distributed analog'' to Turing machines with soft operators (e.g., \texttt{read}, \texttt{write}, and \texttt{address}); its semantics, however, cannot be grounded to actual operations. The latter learns to generate an execution trace in a fully supervised, step-by-step manner.
	
	Another related topic is incorporating neural networks with external (hard) mechanisms. \newcite{harness} propose to better train a neural network by leveraging the classification distribution induced from a rule-based system. \newcite{rationale} propose to induce a sparse code by REINFORCE to make neural networks focus on relevant information. In machine translation, \newcite{supervisedatt} use alignment heuristics to train the attention signal of neural networks in a supervised manner. In these studies, researchers typically leverage external hard mechanisms to improve neural networks' performance.
	
	The uniqueness of our work is to train a fully neuralized/distributed model first, which takes advantage of its differentiability, and then to guide a symbolic model to achieve a meaningful initial policy.
	Further trained by reinforcement learning, the symbolic model's knowledge can improve neural networks' performance by feeding back step-by-step supervision.
	Our work sheds light on neural sequence prediction in general, for example, exploring word alignment~\cite{supervisedatt} or chunking information~\cite{chunk} in machine translation by coupling neural and external mechanisms.

	\section{Conclusion and Future Work}
	In this paper, we have proposed a coupled view of distributed and symbolic execution for natural language queries. By pretraining with intermediate execution results of a distributed executor, we manage to accelerate the symbolic model's REINFORCE training to a large extent. The well-trained symbolic executor could also guide a distributed executor to achieve better performance. Our proposed approach takes advantage of both distributed and symbolic worlds, achieving high interpretability, high execution efficiency, high learning efficiency, as well as high accuracy. 
	
	As a pilot study, our paper raises several key open questions: When do neural networks exhibit symbolic interpretations? How can we better transfer knowledge between distributed and symbolic worlds?
	
	In future work, we would like to design interpretable operators in the distributed model to better couple the two worlds and to further ease the training with REINFORCE. 
	We would also like to explore different ways of transferring knowledge, e.g., distilling knowledge from the action distributions (rather than using the max \textit{a posteriori} action), or sampling actions by following the distributed model's predicted distribution during symbolic one's Monte Carlo policy gradient training (REINFORCE).
	
	\section*{Acknowledgments}
	We thank Pengcheng Yin and Jiatao Gu for helpful discussions; we also thank the reviewers for insightful comments.
	This research is partially supported by the National Basic Research Program of China (the 973 Program) under Grant Nos.~2014CB340301 and 2015CB352201, and the National Natural Science Foundation of China under Grant Nos.~614201091, 61232015 and 61620106007.\bibliography{NE}
	
	\bibliographystyle{icml2017}
\end{document}